\documentclass[runningheads]{llncs}

\usepackage{eccv}

\usepackage{eccvabbrv}

\usepackage{graphicx}
\usepackage{booktabs}

\usepackage[accsupp]{axessibility}  

\usepackage{hyperref}

\usepackage{orcidlink}

\begin{document}

\title{EOL: Transductive Few-Shot Open-Set Recognition by Enhancing Outlier Logits}

\titlerunning{EOL: Transductive Few-Shot Open-Set Recognition}

\author{Mateusz Ochal\inst{1}\and
Massimiliano Patacchiola\inst{2}\and
Malik Boudiaf\inst{3} \and
Sen Wang\inst{4}}

\authorrunning{M.~Ochal et al.}

\institute{Heriot-Watt University, Edinburgh, UK (\email{m.ochal@hw.ac.uk})\footnote{This work was conducted as part of a PhD thesis.} \and
University of Cambridge, UK \and 
Equall.ai, New York, Paris, Lisbon \and 
Imperial College London, London, UK
}

\maketitle

\definecolor{lightgray}{gray}{0.55}
\definecolor{lightsalmon}{rgb}{1.0, 0.63, 0.48}
\definecolor{lightgreen}{rgb}{0.5, 1.0, 0.5}

\newcommand{\std}[1]{\color{lightgray}{\scriptsize{$\pm{#1}$}}}
\newcommand{\knn}{K-NN} 

\newcommand{\rcell}[2]{
    \renewcommand{\arraystretch}{0.5}\specialcell{$#1$\\\std{#2}}}
\newcommand{\rcellbest}[2]{\rcell{\textbf{#1}}{\textbf{#2}}}
\newcommand{\rcellsndbest}[2]{\rcell{\textbf{#1}}{#2}}
\newcommand{\emptycell}{-}
\newcommand{\sidecell}[2]{
    \parbox[t]{1mm}{\multirow{#1}{*}{\rotatebox[origin=c]{90}{#2}}}}

\newcommand{\ourmethod}{EOL}
\newcommand{\ostim}{OSTIM}

\newcommand{\todo}[1]{\textcolor{red!60}{TODO: #1}}

\begin{abstract}
In Few-Shot Learning (FSL), models are trained to recognise unseen objects from a query set, given a few labelled examples from a support set. In standard FSL, models are evaluated on query instances sampled from the same class distribution of the support set. In this work, we explore the more nuanced and practical challenge of Open-Set Few-Shot Recognition (OSFSL). Unlike standard FSL, OSFSL incorporates unknown classes into the query set, thereby requiring the model not only to classify known classes but also to identify outliers. Building on the groundwork laid by previous studies, we define a novel transductive inference technique that leverages the InfoMax principle to exploit the unlabelled query set. We called our approach the Enhanced Outlier Logit (EOL) method. EOL refines class prototype representations through model calibration, effectively balancing the inlier-outlier ratio. This calibration enhances pseudo-label accuracy for the query set and improves the optimisation objective within the transductive inference process. We provide a comprehensive empirical evaluation demonstrating that EOL consistently surpasses traditional methods, recording performance improvements ranging from approximately $+1.3\%$ to $+6.3\%$ across a variety of classification and outlier detection metrics and benchmarks, even in the presence of inlier-outlier imbalance\footnote{The code to reproduce the experiments is available online: \url{https://github.com/mattochal/enhanced-outlier-logits}}.

\end{abstract}

\section{Introduction}
\label{sec:intro}

\newcommand{\softmax}{\mathtt{softmax}}
\newcommand{\sigmoid}{\mathtt{sigmoid}}

\newcommand{\inputSpace}{\mathcal{X}}
\newcommand{\outputSpace}{\mathcal{Y}}

\newcommand{\inputSpaceDef}{\mathcal{X}\in\mathbb{R}^D}

\newcommand{\supportSet}{\mathcal{S}}
\newcommand{\querySet}{\mathcal{Q}_\texttt{IN}}
\newcommand{\totalQuerySet}{\mathcal{Q}}
\newcommand{\outlierQuerySet}{\mathcal{Q}_\texttt{OUT}}
\newcommand{\inoutlierSet}{\hat{\mathcal{Q}}}

\newcommand{\Kquery}{K_{\texttt{IN}}^q}
\newcommand{\Kshot}{K^s}
\newcommand{\Koutlier}{K_{\texttt{OUT}}^q}
\newcommand{\Nway}{{N_{\texttt{IN}}}}

\newcommand{\cardS}{|\supportSet|}
\newcommand{\cardQ}{|\querySet|}
\newcommand{\cardOut}{|\outlierQuerySet|}
\newcommand{\cardTotalQ}{|\totalQuerySet|}

\newcommand{\inlierSpace}{\outputSpace_{\texttt{IN}}}
\newcommand{\outlierSpace}{\outputSpace_{\texttt{OUT}}}

\newcommand{\inlierSpaceDef}{\inlierSpace\!\subseteq\!\outputSpace}
\newcommand{\outlierSpaceDef}{\outlierSpace\!\subseteq\!\outputSpace}

\newcommand{\supportSetX}{\supportSet^x}
\newcommand{\supportSetY}{\supportSet^y}
\newcommand{\querySetX}{\querySet^x}
\newcommand{\querySetY}{\querySet^y}
\newcommand{\outlierQuerySetX}{\outlierQuerySet^x}
\newcommand{\outlierQuerySetY}{\outlierQuerySet^y}
\newcommand{\totalQuerySetX}{\totalQuerySet^x}
\newcommand{\totalQuerySetY}{\totalQuerySet^y}
\newcommand{\inoutlierSetY}{\inoutlierSet^y}

\newcommand{\supportSetDef}{\supportSet\!=\!\{ (x^s_i, y^s_i) \in \inputSpace \times \inlierSpace \}_{i=1}^{\cardS}}
\newcommand{\querySetXDef}{\querySetX\!=\!\{ x^q_i\!\in\!\inputSpace \}_{i=1}^{\cardQ}}
\newcommand{\querySetYDef}{\querySetY\!=\!\{ y^q_i\!\in\!\inlierSpace \}_{i=1}^{\cardQ}}
\newcommand{\outlierQuerySetXDef}{\outlierQuerySetX\!=\!\{ x^q_i\!\in\!\inputSpace \}_{i=1}^{\cardOut}}
\newcommand{\outlierQuerySetYDef}{\outlierQuerySetY\!=\!\{ y^q_i\!\in\!\outlierSpace \}_{i=1}^{\cardOut}}
\newcommand{\totalQuerySetXDef}{\totalQuerySetX\!=\!\querySetX \cup \outlierQuerySetX}
\newcommand{\totalQuerySetYDef}{\totalQuerySetY\!=\!\{ y^q_i\!\in\!\outlierSpace \}_{i=1}^{\cardTotalQ}}
\newcommand{\inoutlierSetYDef}{\inoutlierSetY\!=\!\{\hat{y}_i\}_{i=1}^{|\totalQuerySetX|}}

\newcommand{\fracInline}[2]{#1/#2}

\newcommand{\featureSpace}{\mathcal{Z}}
\newcommand{\normFeatureSpace}{\mathcal{Z}'}
\newcommand{\classOutputSpace}{\mathcal{P}_\texttt{IN}}
\newcommand{\outlierOutputSpace}{\mathcal{P}_\texttt{OUT}}
\newcommand{\backboneW}{\mathbf{\Phi}}
\newcommand{\detectorW}{\mathbf{\Psi}}
\newcommand{\classifierW}{\mathbf{\Omega}}
\newcommand{\backbone}{f_\backboneW}
\newcommand{\detector}{d_\detectorW}
\newcommand{\classifier}{h_\classifierW}
\newcommand{\norm}{\pi}
\newcommand{\backboneDef}{\backbone\!:\!\inputSpace\!\rightarrow\!\featureSpace\!\subset\!\mathbb{R}^D}
\newcommand{\detectorDef}{\detector\!:\!\normFeatureSpace\!\rightarrow\!\outlierOutputSpace\!\subset\!\mathbb{R}}
\newcommand{\classifierDef}{\classifier\!:\!\normFeatureSpace\!\rightarrow\!\classOutputSpace\!\subset\!\mathbb{R}^{\Nway}}
\newcommand{\featureNormDef}{\norm\!:\!\featureSpace\!\rightarrow\!\normFeatureSpace\!\subset\!\mathbb{R}^D}

\newcommand{\projectionFootnote}{\footnote{We use $\supportSet_{x}$ and $\supportSet_{y}$ to denote the first and second projections of $\supportSet$, e.g. $\supportSet_{y}\!=\!\text{proj}_2(\supportSet)\!=\!\{ y^s_i\!\in\!\inlierSpace \}_{i=1}^{\cardS}$}~}

\newcommand{\outlierScore}{o}
\newcommand{\feature}{z}
\newcommand{\classFeatures}{\mathcal{Z}^{\supportSet}_{j}}
\newcommand{\classFeaturesDef}{\classFeatures\!=\!\{ \feature^s_{i} | y^s_i = j \} \subset \featureSpace^\supportSet}
\newcommand{\prototype}{c}
\newcommand{\prototypeDef}{\prototype_j = \sum_{\feature\in\classFeatures} \fracInline{\feature}{\Kshot}}

\newcommand{\logit}{l}
\newcommand{\featureNorm}[1]{\norm(#1)}
\newcommand{\lTwoNorm}[1]{||#1||^2}
\newcommand{\cosineSim}[2]{\langle #1 , #2 \rangle}
\newcommand{\featureNormPlaceholder}{\featureNorm{\cdot}}
\newcommand{\logitDef}{\logit_{ij} = \cosineSim{\featureNorm{\feature_i}}{\featureNorm{\prototype_j}}}
\newcommand{\caliA}{\eta}
\newcommand{\caliB}{\delta}
\newcommand{\logitCalibDef}{\logit_{ij} = \textcolor{red}{\caliA_j} *\cosineSim{\featureNorm{\feature_i}}{\featureNorm{\prototype_j}} + \textcolor{red}{\caliB_j}}

\newcommand{\taskFeatures}{\feature \in \featureSpace^\mathcal{A}}
\newcommand{\taskFeaturesDef}{\featureSpace^\mathcal{A} = \featureSpace^\supportSet \cup \featureSpace^\querySet}
\newcommand{\taskMean}{\mu}
\newcommand{\taskMeanDef}{\mu=\sum_{\taskFeatures} \fracInline{\feature}{(\cardTotalQ+\cardS})}

\newcommand{\Noodway}{N_{\texttt{OUT}}}
\newcommand{\Noodwaycolor}{\textcolor{red}{N_{\texttt{OUT}}}}
\newcommand{\Ntotal}{N_\texttt{IN+OUT}}
\newcommand{\NtotalDefOSTIM}{\Ntotal\!=\!\Nway\!+\!1}
\newcommand{\NtotalDefOURS}{\Ntotal\!=\!\Nway\!+\!\textcolor{red!70}{\Noodway}}

\newcommand{\sumS}{\sum_{i=1}^{|\supportSet|}}
\newcommand{\sumQ}{\sum_{i=1}^{|\querySet|}}
\newcommand{\sumTotalQ}{\sum_{i=1}^{|\totalQuerySet|}}
\newcommand{\sumN}{\sum_{j=1}^{\Nway}}
\newcommand{\sumNk}{\sum_{k=1}^{\Nway}}
\newcommand{\sumNp}{\sum_{j=1}^{\Nway+1}}
\newcommand{\sumNo}{\sum_{j=1}^{\Ntotal}}
\newcommand{\sumO}{\sum_{j=\Nway\!+\!1}^{\Ntotal}}
\newcommand{\iB}{\frac{1}{\beta}}
\newcommand{\iCardQ}[1][1]{\frac{#1}{\cardQ}}
\newcommand{\iCardS}[1][1]{\frac{#1}{\cardS}}
\newcommand{\iNway}[1][1]{\frac{#1}{\Nway}}
\newcommand{\iNtotal}[1][1]{\frac{#1}{\Ntotal}}
\newcommand{\logb}{\log{\iB}}
\newcommand{\cbraq}[1]{ \left \{ #1 \right \}}
\newcommand{\sbraq}[1]{ \left [ #1 \right ]}
\newcommand{\rbraq}[1]{ \left ( #1 \right )}
\newcommand{\logitij}{l_{ij}}
\newcommand{\pqij}{p^{q}_{ij}}
\newcommand{\pij}{p_{ij}}
\newcommand{\pqio}{p^{q}_{i,\Nway\!+\!1}}
\newcommand{\pqioBal}{p^{q}_{i,\Nway\!+\!1}}
\newcommand{\logitijoDef}{l_{i,{\Nway+1}}\!=\!\fracInline{-\sumN \logit_{ij}}{\Nway}}
\newcommand{\pqioDef}{\pqio\!=\!\fracInline{-\sumN \logit_{ij}}{\Nway}}
\newcommand{\pijDef}{\pij=\softmax_j(\logit_{ij})}
\newcommand{\pqioBalDefBal}{\pqioBal\!=\!\fracInline{-\sumN \logit_{ij}}{\Nway}}
\newcommand{\hpqj}{\hat{p}^{q}_{j}}
\newcommand{\hpqo}{\hat{p}^{q}_{\Nway\!+\!1}}
\newcommand{\psij}{p^{s}_{ij}}
\newcommand{\ysij}{y^{s}_{ij}}
\newcommand{\hpqjDef}{\hpqj\!=\!\iCardQ \sumQ \pqij}
\newcommand{\lma}{\lambda^{\texttt{MA}}}
\newcommand{\lco}{\lambda^{\texttt{CO}}}
\newcommand{\lce}{\lambda^{\texttt{CE}}}
\newcommand{\lcemacodef}{\lambda^{\{\texttt{CE},\texttt{MA},\texttt{CO}\}}}
\newcommand{\lreg}{\lambda^{\textsc{REG}}}
\newcommand{\lmaEOLcolor}{\textcolor{red!70}{\lmaEOL}}
\newcommand{\lcoEOLcolor}{\textcolor{red!70}{\lcoEOL}}
\newcommand{\lmaEOL}{{\lambda}_{\texttt{OUT}}^{\texttt{MA}}}
\newcommand{\lcoEOL}{{\lambda}_{\texttt{OUT}}^{\texttt{CO}}}
\newcommand{\maLoss}{\textcolor{teal!90}{\texttt{MA}}}
\newcommand{\coLoss}{\textcolor{blue!75}{\texttt{CO}}}
\newcommand{\ceLoss}{\texttt{CE}}
\newcommand{\maLossEOL}{\textcolor{teal!90}{\mathring{\texttt{MA}}}}
\newcommand{\coLossEOL}{\textcolor{blue!75}{\mathring{\texttt{CO}}}}
\newcommand{\ourMethodLoss}{\texttt{\ourmethod}}
\newcommand{\ostimLoss}{\texttt{\ostim}}

\newcommand{\taskmean}{{\mu_{\texttt{S+Q}}}}

\newcommand{\wo}{w_{\Nway\!+\!1}}
\newcommand{\wj}{w_{j}}

In the context of image classification, Open-Set Recognition (OSR) is a machine learning concept where models encounter \emph{known} (inlier) and \emph{unknown} (outlier) class distributions at test time \cite{geng2020recent}. Their goal is not only to predict the label of inlier samples but also to determine the likelihood that each sample belongs to an outlier category. In the open world, machine learning algorithms may encounter many novel situations that were not seen during training and must be appropriately handled, as is the case in many safety-critical systems \cite{borg2023ergo}. In the context of image recognition, the OSR task aims to study model performance when encountering samples from novel classes. OSR is akin to several related problems, such as outlier/anomaly/out-of-distribution detection \cite{ruff2021unifying}, which often share methodologies \cite{geng2020recent}.

\begin{figure}
    \centering
    \includegraphics[width=\linewidth]{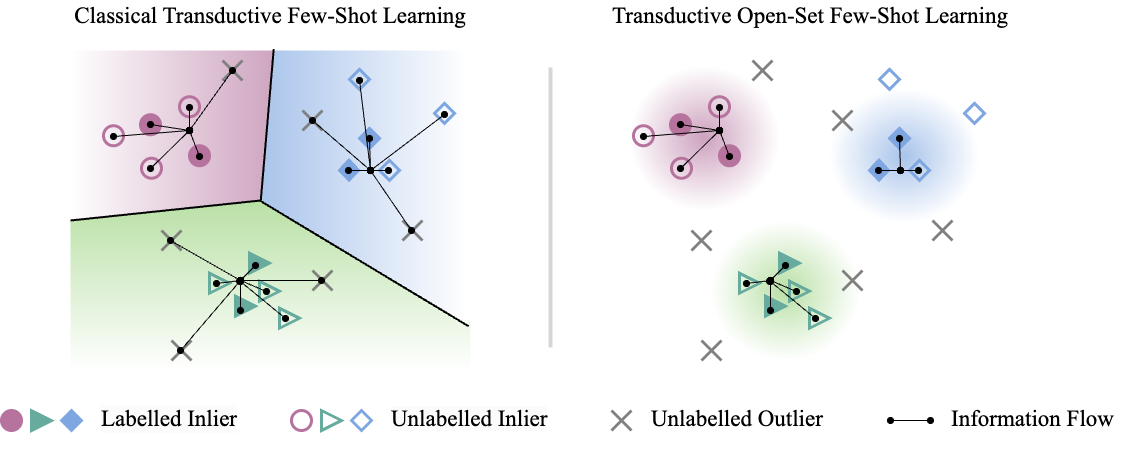}
    \caption[Transductive Open-Set Few-Shot Learning]{In this study, we investigate the Transductive Open-Set Few-Shot Learning setting (right). In contrast to the nominal Transductive Few-Shot Learning task (left), the algorithms need to give special consideration to outlier samples present in the unlabelled query set to restrict the information flow from unwanted outliers during transductive inference.}
    \label{fig:osfsl}
\end{figure}

While much of the literature has concentrated on classical deep learning settings involving medium to large datasets following a long-tail distribution \cite{geng2020recent}, open-set recognition in data-limited settings has received considerably less attention \cite{boudiaf2022ostim,boudiaf2023osem}. Nonetheless, open-set recognition addresses the realistic scenario where a system must recognise objects or patterns that were not seen during its training phase, a critical capability for real-world applications and robotics \cite{geng2020recent,scheirer2012toward}. In light of these challenges, the Open-Set Few-Shot Learning (OSFSL) task was proposed to evaluate Few-Shot Learning (FSL) algorithms in the open-set setting by introducing unknown classes into the query set \cite{boudiaf2022ostim, boudiaf2023osem}. In the restricted view of known classes, represented by a small support set, the OSR challenge becomes even more daunting. Many FSL algorithms lack adequate mechanisms for detecting outliers and need additional consideration \cite{boudiaf2022ostim}. Moreover, many OSR algorithms perform sub-optimally in data-limited settings \cite{boudiaf2022ostim}.

In the FSL literature, transductive inference methods \cite{antoniou2019sca, liu2019learning, dhillon2020baseline, boudiaf2020transductive} have demonstrated notable success in improving performance compared to their entirely inductive counterparts. It has been shown that utilising the query set as an additional unlabelled training set can significantly enhance performance \cite{boudiaf2020transductive} and can also be effectively applied to video recognition \cite{nguyen2022inductive, siam2022temporal} and other applications ranging from natural language processing \cite{colombo2023transductive} to medicine \cite{dai2023pfemed}, and neuroscience \cite{bontonou2021few}. 

However, despite these recent successes, few-shot algorithms are commonly designed to function in a closed-set setting and struggle in dynamic environments with unknown classes that the models may encounter at inference \cite{geng2020recent}. These unknown classes can be viewed as outlier classes from the perspective of the algorithm trained on inlier classes contained in the support set. This inability of algorithms to deal with outlier classes is another hindrance and source of unreliability in real-world scenarios where the classes encountered during inference can be unknown or unbounded \cite{geng2020recent}. The problem of open-set recognition particularly affects transductive algorithms \cite{boudiaf2022ostim, boudiaf2023osem}; as transductive methods rely on information from the evaluation data, any outliers present in the data can taint this information, misleading the model and propagating irrelevant information. Even when outlier classes are expected, the problem of selecting relevant information from unlabelled samples remains a challenge, as the distribution of unknown classes can also be unknown. Having a reliable transductive method capable of handling the OSFSL task in many applications \cite{colombo2023transductive, dai2023pfemed, bontonou2021few} remains a challenge. This forms the motivation to investigate the problem of the transductive FSL task studied in this work.

\begin{figure}[t!]
    \centering
    \includegraphics[width=0.99\linewidth, trim={21 0 2 0}, clip]{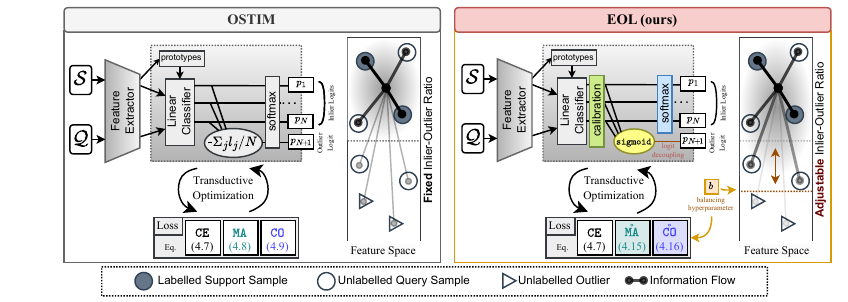}
    \caption[Diagram comparing OSTIM and EOL.]{Diagram comparing OSTIM and our proposed EOL algorithm. Both methods are transductive, utilising the query set as an additional unlabelled training set. \textbf{Left:} OSTIM implicitly assumes a fixed ratio between inlier and outlier classes, which may be easily disrupted in the real world. \textbf{Right:} Key changes introduced in EOL include: (1) decoupling the inlier-outlier representations from the softmax; (2) introducing the balancing hyperparameter $b$ to account for the contribution of inliers and outliers; (3) reformulating and rebalancing the $\maLossEOL$ and $\coLossEOL$ loss terms; and (4) improving prediction confidence through model calibration.}
    \label{ch:openset:fig:ostim_vs_eol}
\end{figure}

Some work has already examined the transductive OSFSL setting \cite{boudiaf2022ostim, boudiaf2023osem}. For example, \cite{boudiaf2022ostim} introduced OSTIM, an open-set variation of the popular Transductive Information Maximisation (TIM) \cite{boudiaf2020transductive}, leveraging the InfoMax principle for transductive inference. However, the algorithm implicitly assumes a fixed ratio between inlier and outlier classes, which may be easily disrupted in the real world. Later, a method called OSLO \cite{boudiaf2023osem} superseded OSTIM in the transductive OSFSL task and alleviated the assumption of known inlier-outlier ratios through an iterative inlier assignment process achieved through a likelihood maximisation algorithm. However, the challenge of reliable few-shot learning in the open-set setting remains. 

In light of these observations, we introduce a novel transductive OSFSL method called Enhanced Outlier Logit (EOL). Our work builds on OSTIM \cite{boudiaf2020transductive} and improves the optimisation objective by decoupling the inlier-outlier representations. Our approach separates and balances the contribution of inliers and outlier loss terms in the transductive loss. In addition, our approach encourages improved prediction confidence through model calibration, which is particularly challenging in data-limited settings. Our proposed solution exhibits remarkable robustness across various types of tasks and consistently demonstrates superior performance in diverse settings. An overview of the differences between OSTIM and EOL is shown in Figure~\ref{ch:openset:fig:ostim_vs_eol}.

Our contributions are as follows:
\begin{enumerate}
    \item We expose the underlying problems over the recently proposed OSTIM \cite{boudiaf2022ostim} method, which leads to sub-optimal performance, as we demonstrate through empirical and theoretical analysis.
    \item We introduce a novel method called EOL, which addresses the limitations of the OSTIM algorithm by appropriately balancing the ratio of known and unknown samples in the open-set setting. 
    \item We demonstrate the effectiveness of our proposed solution through rigorous evaluation involving new tasks with inlier-outlier imbalance, outperforming baselines and previous state-of-the-art solutions in the same category by an average of $+1.3$ to $+6.1$ percentage points.
\end{enumerate}

\section{Background}

Open-set recognition in the FSL setting has received some but limited attention. PEELER \cite{liu2020peeler} leverages the classical prototypes and uses a multivariate Gaussian distribution to represent prototypes. This way, they are able to build a distribution of the classes and learn to distinguish outliers. SnaTCHer \cite{jeong2021snatcher} uses the rule of transformation consistency that assumes similar samples should lie close to each other after performing a transformation. Specifically, their method replaces the prototype with a query sample and hypothesises that transforming it should lead to a similar point in representation space as the transformation of the class prototype if the query also belongs to the class of the same category. Meta-BCE \cite{kozerawski2021metabce} is a one-class classification framework for few-shot learning that learns a separate feature representation for each class. It can be adapted to identify outliers. OOD-MAML \cite{jeong2020ood} extends the classical MAML \cite{finn2017maml} algorithm into the out-of-distribution setting. In this work, the authors use a generator for outlier samples during meta-learning and refine the boundaries of classes within a single task. They meta-learn to model fake pieces using vectors in embedding space via a gradient update in the adversarial direction. FROB \cite{dionelis2021frob} generates outlier samples close to inlier class boundaries to find the class distributions in a self-supervised way. Willes et al. \cite{willes2022bayesian} propose few-shot learning for an open-world recognition framework (FLOWR), which differs from the standard task in the amount of relevant context data. Their framework offers a way of continuously detecting and learning novel classes during operation. Related works also mention out-of-distribution samples \cite{le2021poodle,ren2018kmeans}. However, here, the task definition is slightly different and serves to improve the standard FSL task rather than to augment the method to the Open-Set Few-Shot Recognition task. POODLE \cite{le2021poodle} uses samples outside the context of the standard FSL task to refine the boundaries of inlier classes to benefit classification accuracy. Semi-supervised FSL \cite{ren2018kmeans} used a catch-all cluster for `distractor' classes present in the unlabelled data given with the support set in the context of semi-supervised learning. 

\section{Methodology}
\label{ch:openset:sec:method}

\subsection{Task Definitions}

\subsubsection{Standard Task} 
In image classification, a typical FSL task is defined by a small, labelled \emph{support set} and an unlabelled \emph{query set}. Formally, given an input space $\inputSpace$ and a target space $\outputSpace$, the support set $\supportSetDef$ contains $\cardS$ input-target pairs over a closed set (\emph{inlier}) classes $\inlierSpaceDef$. The algorithm aims to correctly predict labels for the query samples $\querySetXDef$ with corresponding ground-truth labels $\querySetYDef$, which are unavailable during testing. The standard $\Kshot$-shot $\Nway$-way setting assumes class-balanced labels, where for every one of $\Nway\!=\!|\inlierSpace|$ unique classes in the given task, there are exactly $\Kshot$ and $\Kquery$ unique samples in the support and query set, respectively, i.e., $\cardS\!=\!\Kshot\!\times\!\Nway$ and $\cardQ\!=\!\Kquery\!\times\!\Nway$.

\subsubsection{Open-Set Task Variation}
The OSFSL setting augments the standard FSL task by injecting a set of \emph{outlier} queries. Specifically, an additional set of samples $\outlierQuerySetXDef$ are drawn from a separate set of classes $\outlierQuerySetYDef$, where $\outlierSpace\!\subseteq\!\outputSpace$, and the classes are non-overlapping with the support set, i.e. $\inlierSpace\cap\outlierSpace\!=\!\emptyset$. In this study, $|\outlierSpace|=\!\Koutlier\!\times\!\Nway$, where $\Koutlier$ is the number of samples per outlier class. The outlier set and inlier set are mixed together to form a single input set $\totalQuerySetXDef$, with binary labels $\inoutlierSetYDef$, where:
\begin{align}
    \hat{y}_i = \left\{
        \begin{array}{ll}
            -1 & \mbox{if  } y_i \in \querySetY \\
            +1 & \mbox{if  } y_i \in \outlierQuerySetY
        \end{array}
    \right.
\end{align}

Given the labelled support set $\supportSet$ and the unlabelled query set $\totalQuerySetX$, the goal of a model is to correctly perform two tasks: (1) a binary classification task over $\inoutlierSetY$, predicting whether a query sample $x^q_i\in\totalQuerySetX$ is an inlier or an outlier; and (2) an $\Nway$-way classification task to correctly predict the inlier label in $\querySetY$ of the inlier samples in $\querySetX$.

\subsection{Preliminaries}

Following the experiments in previous work \cite{boudiaf2023osem,boudiaf2022ostim}, we keep a common structure between all the methods tried in our evaluation. Specifically, we separate the feature extractor model from the classification model and perform an additional task-adaptation procedure. This setup allows us to focus our evaluation on outlier detection and multi-class classification, regardless of the complexity of the backbone. 

\textbf{Feature extraction:} Formally, for a given FSL evaluation task, we assume a pre-trained feature extractor, $\backboneDef$, that transforms the input space $\inputSpace$ into a feature space $\featureSpace$ and is shared between all tasks. The model weights $\backboneW$ are pre-trained on a large available dataset and frozen during all other stages of evaluation. We can perform a feature-adaptation procedure $\featureNormDef$ to further transform the features conditioned on the given task.

\textbf{Outlier detector:} The adapted features are then classified using a binary classifier $\detectorDef$ which takes a threshold and a feature $\feature_i\!\in\!\normFeatureSpace$ as input and generates a score $\outlierScore_i$ representing the likelihood of the sample being an outlier. An appropriate threshold value can be selected over $\outlierScore_i$ to balance the precision and recall of the detector depending on the application.

\textbf{Multi-class classifier:} In parallel, a multi-class classifier $\classifierDef$ maps features $\feature_i$ to the posterior distribution over $\Nway$ unique labels. The weights of the combined model $(\backboneW,\detectorW,\classifierW)$ could be learned via classical transfer-learning \cite{chen2019closer} approaches or using meta-learning \cite{ziko2020laplacian,snell2017proto,sung2017relationnet} or a combination of both \cite{triantafillou2019meta,triantafillou2021universal}. 

\subsection{Original OSTIM}
The OSTIM~ \cite{boudiaf2022ostim} was first introduced to address the challenges of open-set recognition for transductive few-shot learning methods, building on the success of the original transductive inference method called TIM \cite{boudiaf2020transductive} for the closed-set setting. At its core, the closed-set algorithm uses class prototypes \cite{snell2017proto,wang2019simpleshot} and adjusts their location in the feature space using transductive inference to incorporate information using the query set as an additional unlabelled set of points. The open-set variation accounts for potential outliers by representing the outlier logit using the negative sum of the class logits.

\textbf{Feature norm}. Formally, as the task adaptation step, features are normalised by subtracting the mean of all features in a given task: $\featureNorm{z}=z-\taskmean$, where the mean feature is defined as:
\begin{equation}
    \taskmean = \frac{1}{\cardS + \cardTotalQ} ( \sumS z^s_i + \sumTotalQ z^q_i)
\end{equation}

Then, cosine similarity is used to compare the distances between the normalised vectors and the normalised class centroid (prototypes) to work out the likelihood of the sample belonging to one of the inlier classes. The prototype of the inlier class $j$ ($1 \leq j\leq\Nway$) with features in $\classFeaturesDef$ is calculated as $\prototypeDef$. The logit $\logitij$ of the $i$-th sample and the $j$-th class can then be obtained by measuring cosine distance $\cosineSim{\cdot}{\cdot}$ between them:
\begin{equation}\label{ch:openset:eq:cosine}
    \logitDef
\end{equation}
Thus, we can obtain the probability that a sample $i\in|\featureSpace^\mathcal{A}|$ belongs to class $j\in[1, \dots, \Nway]$ by taking the $\softmax$ over all classes: 
\begin{equation}\label{ch:openset:eq:ostim prob}
    \pijDef
\end{equation}

\textbf{Outlier logit}. The authors construct an outlier logit representing the distance to a `catch-all' outlier class $\Nway+1$. They define this logit as the negative mean of inlier logits: 
\begin{equation}\label{ch:openset:eq:ostim outlier logit}
    \logitijoDef
\end{equation}
In a simplified intuition, this outlier logit indicates how dissimilar the sample is to the inliers, representing the general space that lies orthogonal to the mean of the inlier prototypes.

\textbf{OSTIM Loss}. The OSTIM uses principles from mutual information to derive a loss for optimising the location of the class centroids while using the query set as an additional unlabelled set of data points. The loss is a weighted sum of three terms: cross-entropy ($\ceLoss$), marginal entropy ($\maLoss$), and conditional entropy ($\coLoss$):
\begin{align}
    L_{\ostimLoss} &= \iCardS[\lce] \cdot \ceLoss + \lma \cdot \maLoss - \iCardQ[\lco] \cdot \coLoss \\
    \ceLoss &= \sumS \sumN \ysij \log \psij \\
    \maLoss &= \sumNo \hpqj \log \hpqj \\
    \coLoss &= \sumQ \sumNo \pqij \log \pqij
\end{align}
The loss is weighted by three hyperparameters $\lcemacodef>0$, which are found through validation on a held-out dataset. The $\ysij$ denotes one-hot positional encoding of labels $y^s_i \in \supportSet_{y}$. The output logits are defined as $\hpqjDef$, and $\NtotalDefOSTIM$. The effect of the $\lco$ loss is similar to assigning pseudo-labels to the query set samples to drive the optimisation objective in the direction that reaffirms confident predictions. On the other hand, $\lma$ has the counter effect of spreading the confidence of predictions equally across all classes, which can act as a regularisation for the objective function.

\subsection{Limitations of OSTIM}\label{ch:openset:sec:ostim limitations}
The original OSTIM method presents a noteworthy improvement compared to classical closed-set algorithms and other transductive algorithms within the same category. However, we delve into the intricacies of the existing method and identify the key limitations and areas for improvement. Our exploration revolves around three key points, each shedding light on the unique aspects of the original approach.

\subsubsection{1) Analysing the Inlier-Outlier Imbalance Problem}
The first challenge arises in how the original method treats outlier and inlier classes. Incorporating the outlier class within the softmax formulation introduces an intrinsic imbalance issue. To illustrate, in the context of a conventional task comprising 15 query samples, with 5 inlier classes and 5 outlier classes, the consequence is that the `catch-all' outlier class will be associated with 75 query outlier samples (since all outlier samples are associated with a single output) whereas the remaining inlier classes will each be allocated a mere 15 samples. Different ratios of the inlier and outlier samples will, therefore, have varying degrees of imbalance issues. We refer to this problem as the `Inlier-Outlier Imbalance'.

The Inlier-Outlier Imbalance problem affects OSTIM's optimisation. We can decompose the original formulation of OSTIM's losses $\maLoss$ and $\coLoss$ (defined in~\ref{ch:openset:eq:lma_expanded} and \ref{ch:openset:eq:lco_expanded}) into two separate components for the inlier and outlier classes:
\begin{align}
    \maLoss &=
    \underbrace{\sumN \hpqj \log \hpqj }_{\text{term for inliers}} \hspace{1.5em} + \hspace{1.5em}
    \underbrace{\mathop{\phantom{\sumN}} \hspace{-1.5em} \hpqo \log \hpqo}_{\text{term for outliers}} \label{ch:openset:eq:lma_expanded}
    \\
    \coLoss &=
    \underbrace{\sumQ ( \sumN \pqij \log \pqij}_{\text{term for inliers}} \hspace{1.5em} + \hspace{1.5em}
    \underbrace{\mathop{\phantom{\sumN}} \hspace{-1.5em} \pqio \log \pqio}_{\text{term for outliers}} )
    \label{ch:openset:eq:lco_expanded}
\end{align}
This expanded formula remains equivalent to the original OSTIM \cite{boudiaf2022ostim}. The separation shows that inlier and outlier terms are equally weighted, irrespective of their true ratio. This mismatch can cause suboptimal performance.

\subsubsection{2) Analysing the Role of the Outlier Logit}
Furthermore, the formulations in~\ref{ch:openset:eq:ostim prob} can be interpreted as (uncalibrated) probabilities for $P(y^{+1}| x,\supportSet,\totalQuerySetX)$, where the outlier class is treated as one of the valid labels for $y$, i.e. $y^{+1}\in\{1,...,N,N+1\}$. This yields a more complicated optimisation problem as the quality and scale of the outlier logit estimation will have a direct impact on the softmax probabilities for the inlier classes.


\subsubsection{3) Understanding Model Calibration}
Training on a small number of data points leads to high variance and bias, and this is particularly problematic in the FSL setting, resulting in poor model calibration. Model calibration and its effect on the performance of semi-supervised algorithms have been widely studied \cite{loh2022importance} in the general community. Model calibration \cite{guo17calibration} measures how the extent to which a model's output prediction reflects its predictive uncertainty. Given that OSTIM uses pseudo-labels (i.e. the model's predictions) over the query set to drive the optimisation objective, the calibration of the model becomes an important aspect to consider in the transductive FSL setting. Intuitively, a poorly calibrated model could lead to overconfident predictions and bias in the optimisation process, leading to sub-optimal performance of the algorithm.

\subsection{Enhancing the Outlier Logit: Introducing \ourmethod}

\subsubsection{Decoupling the Outlier Logit}
To help us solve the issues above, let us start by decoupling the outlier logit from the complicated influence softmax, thereby allowing us more control over the representation of outliers. Whereas the formulations in~\ref{ch:openset:eq:ostim prob} can be interpreted as $P(y^{+1}| x,\supportSet,\totalQuerySetX)$, an alternative approach is to decouple the inlier likelihoods by conditioning on the probability of selecting an inlier \cite{boudiaf2023osem,willes2022bayesian}, i.e.:
\begin{equation}\label{ch:openset:eq:prob decoupled}
    P(y \mid x, \supportSet, \totalQuerySetX) \propto P(y \mid x, \supportSet, \totalQuerySetX, \hat{y}_i = -1) \times P(\hat{y} = -1 \mid x, \supportSet, \totalQuerySetX)
\end{equation}
Recall that $\hat{y}_i = -1$ indicates an inlier, whereas $\hat{y}_i = +1$ is an outlier. We can assume that the probability of a sample belonging to a particular inlier class given that the sample is an outlier is zero, i.e. $P(y \mid x, \supportSet, \totalQuerySetX, \hat{y}_i = +1) = 0$. Thus, we can perform the softmax function from~\ref{ch:openset:eq:ostim prob} over the inlier classes only, i.e. $j$ ($1 \leq j\leq\Nway$), and with it represent $P(y \mid x, \supportSet, \totalQuerySetX, \hat{y}_i = -1)$, while tackling the probabilities $P(\hat{y} = -1 \mid x, \supportSet, \totalQuerySetX)$ separately.

\subsubsection{Representing the Inlier-Outlier Scores}
OSTIM \cite{boudiaf2022ostim} represents the outlier logits as the negative mean of the inlier logits before applying the softmax function. OSLO \cite{boudiaf2023osem} employs a likelihood maximisation algorithm to compute `inlierness' scores for each sample. In contrast, EOL calculates the inlier score using a $\sigmoid$ function of the logarithm of the exponential sum of the inlier logits. The inlier logit for the $i$-th sample is defined as:
\begin{equation}\label{ch:openset:eq:inlier logit}
    P(\hat{y}=-1 \mid x, \supportSet, \totalQuerySetX) = \sigmoid \left ( - \log\left( \sumN e^{\logitij}\right) + \log(\Nway) -\log(b) \right )
    \end{equation}
where $b$ is a balancing hyperparameter that will be introduced next. The $P(\hat{y}=+1 \mid x, \supportSet, \totalQuerySetX)$ is simply the complementing probability, such that the sum of the probabilities equals 1.

\subsubsection{Addressing Inlier-Outlier Imbalance}
To address the challenge of inlier-outlier imbalance in transductive learning models, we introduce a novel approach through the integration of a balancing hyperparameter, denoted as $b$. This hyperparameter, with values ranging from 0 to 1, is linked to the prior probability $P(\hat{y})$. We adjust the loss based on the anticipated ratio of inliers to outliers within the query set, thereby helping to mitigate the inlier-outlier imbalance problem. Formally, we define the weight $w$ for a given inlier class $j$ and the outlier $\Nway\!+\!1$ is computed as follows: 
\begin{equation}
    \fbox{$\displaystyle \wj = \frac{\Nway}{(1 - b)}$} \hspace{5em}
    \fbox{$\displaystyle \wo = \frac{1}{b} $}\label{ch:openset:eq:weights}
\end{equation}
Thus, we redefine $\maLoss$ loss defined in (\ref{ch:openset:eq:lco_expanded}) and weigh the classes as follows:
\begin{equation}
    \maLossEOL = \sumN \wj \hpqj \log \hpqj + \wo \hpqo \log \hpqo \label{ch:openset:eq:lma_eol}
\end{equation}
where we define $ \hpqo = \sumTotalQ (1 - \sumN \pqij) / \cardTotalQ$. In addition, we redefine $\coLoss$ loss from (\ref{ch:openset:eq:lma_expanded}) to only focus on the inlier classes:
\begin{equation}
    \coLossEOL = \sumQ \sumN \pqij \log \pqij \label{ch:openset:eq:lco_eol}
\end{equation}


\subsubsection{Calibrating Confidence}
In Section~\ref{ch:openset:sec:ostim limitations}, we highlighted the challenge of model calibration in semi-supervised learning. Our approach integrates a conventional model calibration technique inspired by Platt Scaling, which allows dynamic adjustment of the scaling and shifting parameters during transductive optimisation. We found that the re-calibration of inlier logits helped stabilise the model and improve the overall performance. Specifically, we scale and shift the inlier logits for each class $j$ with the following formula:
\begin{equation}\label{ch:openset:eq:cosine_calib}
    \logitCalibDef
\end{equation}
where $\caliA, \caliB \in \mathbb{R}^\Nway$ are parameters optimized during transductive inference.

\subsubsection{Unveiling \ourmethod}
An intuitive diagram of the EOL algorithms is depicted in Figure~\ref{ch:openset:fig:ostim_vs_eol}. To recap, we decouple the Outlier Logit from OSTIM's softmax representation, thereby allowing us more control over the representation of outliers in (\ref{ch:openset:eq:prob decoupled}). We proposed a new way of representing the inlier scores through the sigmoid function in (\ref{ch:openset:eq:inlier logit}). We addressed the inlier-outlier imbalance challenge by implementing a balancing hyperparameter to appropriately address the inlier and outlier contributions in the loss function in (\ref{ch:openset:eq:lma_eol}) and (\ref{ch:openset:eq:lco_eol}). Lastly, we allowed model calibration parameters, defined in (\ref{ch:openset:eq:cosine_calib}), to be optimised through the transductive inference. Finally, after putting everything together, we formulate our $\ourMethodLoss$ loss:
\begin{equation}
    \colorlet{oldcolor}{.}
    \color{black!60}
    \boxed{\color{oldcolor}L_{EOL} = \iCardS[\lce] \cdot \ceLoss + \iNtotal[\lma] \cdot \maLossEOL - \iCardQ[\lco] \cdot \coLossEOL}
\end{equation}

\section{Experiments}
\label{ch:openset:sec:exp}

\subsection{Setup}
\textbf{Implementation.} Our implementation is based on the official open-sourced code repository for OSTIM/OSLO algorithms \cite{boudiaf2022ostim,boudiaf2023osem}, which already implements a significant number of inductive and transductive methods for open-set and closed-set recognition. We make a comprehensive comparison spanning different outlier detection and Few-Shot Learning (FSL) paradigms evaluated in the Open Set Few-Shot Learning (OSFSL) setting. Specifically, we use typical Out-of-Distribution (OOD) detection methods such as \knn~($k$-Nearest Neighbors), which classifies new data points based on the majority label of the \(k\) closest points \cite{altman1992introduction}; IForest (Isolation Forest), that detects anomalies by isolating outliers using tree structures \cite{liu2008isolation}; OCSVM (One-Class Support Vector Machine), which learns a decision function for regions in the feature space treating points outside these regions as outliers \cite{scholkopf2001estimating}; PCA (Principal Component Analysis), that uses distances from the data points to the subspace for detecting outliers \cite{jolliffe2016principal}; and COPOD (Copula-Based Outlier Detection), which estimates outlier scores using empirical copulas for multivariate data \cite{li2020copod}, are included in our evaluation. Inductive FSL algorithms (SimpleShot \cite{wang2019simpleshot}, Baseline\texttt{++} \cite{chen2019closerfewshot}, FEAT \cite{ye2020few}, PROSPER \cite{zhou2021learning}) are paired with a MaxProb (Maximum Probability) algorithm for outlier detection. Transductive FSL approaches (LaplacianShot \cite{ziko2020laplacian}, BDCSPN \cite{liu2020prototype}, TIM-GD \cite{boudiaf2020transductive}, PT-MAP \cite{hu2021leveraging}) are also paired with MaxProb for outlier detection. For the transductive OSFSL setting, we compare our algorithm against OSTIM \cite{boudiaf2022ostim} and OSLO \cite{boudiaf2023osem} algorithms. For the main experiments, we use a pretrained ResNet-12 on the popular ImageNet dataset for feature extraction, which remains fixed throughout all experiments. We include additional experiments in the supplementary material. 

\textbf{Benchmark.} The performance of algorithms is evaluated on the popular MiniImageNet dataset \cite{vinyals2016matching} using the open-set 5-shot 5-way task with 5 outlier classes. Uniquely, we introduce inlier-outlier into the evaluation protocol of OSFSL. Specifically, we vary the ratio of the inlier and outlier samples using $\Kquery$ and $\Koutlier$ such that the number of query samples between tasks remains fixed ($\cardTotalQ=150$). Following previous work \cite{boudiaf2023osem,boudiaf2022ostim}, we report typical metrics for classification and outlier detection averaged over 1000 tasks, namely: accuracy (Acc.), area under the receiver operating characteristic curve (AUROC), area under the precision-recall curve (AUPR), and precision at 90\% recall (Prec@0.9). AUROC is a widely accepted metric for evaluating classification performance under varying thresholds, as it provides a balanced measure of sensitivity and specificity \cite{fawcett2006introduction}. AUPR is particularly useful in scenarios with class imbalance, as it focuses on the performance of the positive class and provides insights into precision-recall trade-offs \cite{davis2006relationship}. Precision at 90\% recall (Prec@0.9) is critical in open-set recognition to ensure that the model maintains high precision when the recall is high, addressing the need for reliability in detecting outliers among a high number of inliers. We repeat each experiment 7 times.

\subsection{Standard Balanced Setting}
In Table~\ref{ch:openset:tab:benchmark_results_5shot}, our selected methods are compared using the standard 5-shot 5-way OSFSL setting with 15 queries for each inlier and outlier class ($\Kquery$=$\Koutlier$=15). 

\textbf{Key Observation}. Our EOL algorithm achieves the highest scores across all metrics, highlighting its effectiveness in both classification accuracy and across the outlier detection metrics. EOL achieves an advantage of at least $+1.3$ to $+35.4$ points across all four metrics compared to all other methods. In comparison to OSTIM \cite{boudiaf2022ostim}, EOL offers a $+2.0$ point advantage in accuracy and achieves a $+2.8$ to $+5.4$ point advantage across other metrics. In comparison to OSLO \cite{boudiaf2023osem}, EOL offers a $+1.3$ point advantage in accuracy, and achieves a $+4.3$ to $+6.1$ point advantage across other metrics. Overall, EOL provides a strong baseline across all categories.

\textbf{Additional Observations}. 
The performances of transductive open-set approaches (OSFSL-T) remain consistently higher compared with other groups, showcasing the strength of this methodology. Compared with the inductive group counterpart (OSFSL), the methods achieve $+7.9$ to $+8.4$ average point advantage across AUPR, AUROC, and Prec@90 metrics, respectively, while also achieving $+2.5$ additional points in accuracy on average. Comparing EOL with the closed-set FSL group (FSL and FSL-T), our method achieves $+3.1$ point improvement in accuracy and a whopping $+17.7$ to $+23.0$ improvement across the other metrics. This highlights the strength of EOL but also highlights the need for continued exploration of classical FSL methods in the open-set setting. Among the traditional approaches, \knn~remains the highest scoring, surpassing the MaxProb paired with FSL and FSL-T approaches, suggesting that \knn~would be a compelling outlier detection method to be integrated with the FSL and FSL-T methods, as observed in \cite{boudiaf2023osem}. 

\begin{table}[t!]
    \centering
    \small
    \caption[Standard Balanced Open-Set Few-Shot Task with 5-shots.]{\textbf{Standard Balanced Open-Set Few-Shot Task.} Results were obtained on 5-shot tasks with all methods using ResNet-12 as a backbone, averaged over 7 initialisation seeds, and evaluated on 1000 tasks each (7000 tasks in total for each method). Bold font represents the best result. The evaluation shows performance for methods from their representative fields: classical out-of-distribution detection methods (OOD), classical few-shot learning (FSL), open-set few-shot learning (OSFSL), and their transductive setting counterparts (-T).}
    \label{ch:openset:tab:benchmark_results_5shot}

\scalebox{1.0}{
       \begin{tabular}{cccccc}
              \toprule
                                                        Method &                              Paradigm &                   Acc. &                  AUROC &                   AUPR &               Prec@0.9 \\
              \midrule
                                      \knn \cite{knn_detector} &     \color{gray!50}{\scriptsize{OOD}} &                      - &          76.2\std{0.6} &          76.1\std{0.6} &          61.4\std{0.5} \\
                               IForest \cite{iforest_detector} &     \color{gray!50}{\scriptsize{OOD}} &                      - &          63.3\std{0.6} &          62.0\std{0.6} &          55.0\std{0.3} \\
                                   OCVSM \cite{ocsvm_detector} &     \color{gray!50}{\scriptsize{OOD}} &                      - &          68.8\std{0.7} &          66.0\std{0.7} &          59.2\std{0.4} \\
                                       PCA \cite{pca_detector} &     \color{gray!50}{\scriptsize{OOD}} &                      - &          75.3\std{0.6} &          75.4\std{0.6} &          60.7\std{0.5} \\
                                   COPOD \cite{copod_detector} &     \color{gray!50}{\scriptsize{OOD}} &                      - &          51.8\std{0.6} &          52.8\std{0.5} &          51.2\std{0.2} \\
              \midrule
                SimpleShot \cite{wang2019simpleshot}  &     \color{blue!50}{\scriptsize{FSL}} &          81.5\std{0.4} &          70.6\std{0.6} &          69.9\std{0.6} &          58.0\std{0.4} \\
               Finetune \cite{chen2019closerfewshot}  &     \color{blue!50}{\scriptsize{FSL}} &          81.5\std{0.4} &          66.3\std{0.6} &          65.4\std{0.5} &          56.5\std{0.3} \\
              \midrule
              LaplacianShot \cite{ziko2020laplacian}  & \color{orange!50}{\scriptsize{FSL-T}} &          82.8\std{0.4} &          57.1\std{0.8} &          57.8\std{0.6} &          52.6\std{0.3} \\
                      BDCSPN \cite{liu2020prototype}  & \color{orange!50}{\scriptsize{FSL-T}} &          82.5\std{0.4} &          61.2\std{0.7} &          62.2\std{0.6} &          53.2\std{0.3} \\
               TIM-GD \cite{boudiaf2020transductive}  & \color{orange!50}{\scriptsize{FSL-T}} &          82.4\std{0.4} &          67.1\std{0.5} &          65.8\std{0.5} &          56.7\std{0.3} \\
                      PT-MAP \cite{hu2021leveraging}  & \color{orange!50}{\scriptsize{FSL-T}} &          78.0\std{0.5} &          62.9\std{0.6} &          62.6\std{0.6} &          54.5\std{0.3} \\
              \midrule
                             OpenMax \cite{bendale2016towards} &  \color{green!50}{\scriptsize{OSFSL}} &          82.0\std{0.4} &          77.3\std{0.6} &          77.3\std{0.6} &          62.1\std{0.5} \\
                                PROSER \cite{zhou2021learning} &  \color{green!50}{\scriptsize{OSFSL}} &          80.0\std{0.4} &          74.9\std{0.6} &          75.3\std{0.6} &          60.1\std{0.5} \\
              \midrule
                                 OSTIM \cite{boudiaf2022ostim} &  \color{red!50}{\scriptsize{OSFSL-T}} &          82.6\std{0.4} &          83.8\std{0.4} &          84.0\std{0.4} &          67.5\std{0.6} \\
                                   OSLO \cite{boudiaf2023osem} &  \color{red!50}{\scriptsize{OSFSL-T}} &          83.3\std{0.4} &          82.6\std{0.4} &          82.5\std{0.5} &          66.8\std{0.6} \\
                                             \ourmethod (ours) &  \color{red!50}{\scriptsize{OSFSL-T}} & \textbf{84.6\std{0.4}} & \textbf{87.2\std{0.4}} & \textbf{86.8\std{0.4}} & \textbf{72.9\std{0.6}} \\
              \bottomrule
              \end{tabular}
              
}

\end{table}

\subsection{Inlier-Outlier Imbalance Results}
In Table~\ref{ch:openset:tab:benchmark_results_5shot_imbalance}, the performances of various methods are evaluated in the imbalanced evaluation task. The results represent averages of 21,000 tasks sampled with three different inlier-outlier imbalance configurations (uniform, low, and high inlier-outlier ratio). This setting is motivated by real-world scenarios where class distribution is typically encountered.

\begin{table}[t!]
    \centering
    \small
    \caption[Imbalanced Open-Set Few-Shot Task with 5-shots using ResNet-12 as backbone.]{\textbf{Imbalanced Open-Set Few-Shot Task with 5-shots using ResNet-12 backbone.} Results averaged over 7 initialisation seeds evaluated on 3000 imbalanced tasks each. Bold font represents the best result. The evaluation shows performance for methods from their representative fields: classical out-of-distribution detection methods (OOD), classical few-shot learning (FSL), open-set few-shot learning (OSFSL), and their transductive setting counterparts (-T).}
    \label{ch:openset:tab:benchmark_results_5shot_imbalance}
    \scalebox{1.0}{

\begin{tabular}{cccccc}
\toprule
                                          Method &                              Paradigm &                   Acc. &                  AUROC &                   AUPR &               Prec@0.9 \\
\midrule
                            \knn \cite{knn_detector} &     \color{gray!50}{\scriptsize{OOD}} &                      - &          76.2\std{0.6} &          72.0\std{0.6} &          58.6\std{0.4} \\
                     IForest \cite{iforest_detector} &     \color{gray!50}{\scriptsize{OOD}} &                      - &          63.3\std{0.6} &          59.4\std{0.5} &          53.7\std{0.2} \\
                     OCVSM \cite{ocsvm_detector} &     \color{gray!50}{\scriptsize{OOD}} &                      - &          68.9\std{0.7} &          62.9\std{0.6} &          56.9\std{0.4} \\
                            PCA &     \color{gray!50}{\scriptsize{OOD}} &                      - &          75.3\std{0.6} &          71.4\std{0.6} &          58.0\std{0.4} \\
                     COPOD \cite{copod_detector} &     \color{gray!50}{\scriptsize{OOD}} &                      - &          52.2\std{0.7} &          53.3\std{0.5} &          51.4\std{0.2} \\
\midrule
       SimpleShot \cite{wang2019simpleshot}  &     \color{blue!50}{\scriptsize{FSL}} &          81.6\std{0.5} &          70.5\std{0.6} &          65.7\std{0.5} &          55.9\std{0.3} \\
       Finetune \cite{chen2019closerfewshot}  &     \color{blue!50}{\scriptsize{FSL}} &          81.7\std{0.5} &          66.4\std{0.6} &          62.0\std{0.4} &          54.8\std{0.2} \\
\midrule
LaplacianShot \cite{ziko2020laplacian}  & \color{orange!50}{\scriptsize{FSL-T}} &          82.3\std{0.5} &          57.8\std{0.7} &          57.4\std{0.5} &          52.6\std{0.2} \\
              BDCSPN \cite{liu2020prototype}  & \color{orange!50}{\scriptsize{FSL-T}} &          82.1\std{0.5} &          60.8\std{0.7} &          59.9\std{0.5} &          52.9\std{0.3} \\
       TIM-GD \cite{boudiaf2020transductive}  & \color{orange!50}{\scriptsize{FSL-T}} &          82.5\std{0.5} &          67.1\std{0.6} &          62.0\std{0.4} &          54.9\std{0.2} \\
              PT-MAP \cite{hu2021leveraging}  & \color{orange!50}{\scriptsize{FSL-T}} &          77.9\std{0.5} &          62.6\std{0.7} &          59.8\std{0.5} &          53.4\std{0.3} \\
\midrule
                     OpenMax \cite{bendale2016towards} &  \color{green!50}{\scriptsize{OSFSL}} &          82.0\std{0.5} &          77.3\std{0.6} &          73.2\std{0.6} &          59.1\std{0.4} \\
                     PROSER \cite{zhou2021learning} &  \color{green!50}{\scriptsize{OSFSL}} &          80.0\std{0.5} &          74.9\std{0.6} &          71.3\std{0.6} &          57.6\std{0.4} \\
\midrule
                     OSTIM \cite{boudiaf2022ostim} &  \color{red!50}{\scriptsize{OSFSL-T}} &          82.4\std{0.5} &          82.0\std{0.5} &          75.0\std{0.5} &          61.9\std{0.4} \\
                     OSLO \cite{boudiaf2023osem} &  \color{red!50}{\scriptsize{OSFSL-T}} &          82.6\std{0.5} &          80.6\std{0.5} &          73.6\std{0.5} &          61.3\std{0.4} \\
                                   \ourmethod (ours) &  \color{red!50}{\scriptsize{OSFSL-T}} & \textbf{84.3\std{0.4}} & \textbf{84.5\std{0.4}} & \textbf{76.6\std{0.5}} & \textbf{64.3\std{0.4}} \\
\bottomrule
\end{tabular}

}
\end{table}

\textbf{Key Observations.} Most methods experience a marginal drop in performance in the imbalanced setting compared to the standard balanced tasks, highlighting the challenges of this setting. Nonetheless, our EOL algorithm remains the highest-achieving method across all metrics, demonstrating its robustness in the imbalanced setting. EOL achieves a $+1.6$ to $+32.3$ point advantage across all four metrics compared to all other algorithms. EOL offers a $+1.9$ point advantage in accuracy over the OSTIM algorithm, while achieving $+1.6$ to $+2.5$ point advantage in outlier detection capability. In comparison to OSLO, EOL yields a $+1.7$ point advantage in accuracy, while producing $+3.0$ to $+3.9$ point advantage in outlier detection capability. 

\textbf{Additional Observations.} In the imbalanced settings, most methods experience a slight drop in performance compared to the balanced scenario in Table~\ref{ch:openset:tab:benchmark_results_5shot}. The largest drops are observed in the transductive methods (FSL-T and OSFSL-T), which is expected since these methods heavily rely on the quality of the query set. OSFSL-T dropped by about $-2.2$, $-9.3$, and $-6.5$ points in AUROC, AUPR, and Prec@0.9 - highlighting the challenges associated with the imbalanced setting. In comparison, FSL-T algorithms drop only in AUPR and Prec@90 with reported drops of $2.3$ and $0.8$, respectively. This further highlights the sensitivity of OSFSL-T methods on imbalance. Other methods collectively (i.e. OOD, FSL, OSFSL) report similar performance drops of $3.2$ and $2.2$ points in AUPR and Prec@0.9. AUROC and accuracy are the metrics that are least affected by imbalance. Interestingly, LaplacianShot \cite{ziko2020laplacian} displays the highest robustness to the imbalance.

\begin{figure}[b!]
    \centering
    \includegraphics[width=\linewidth]{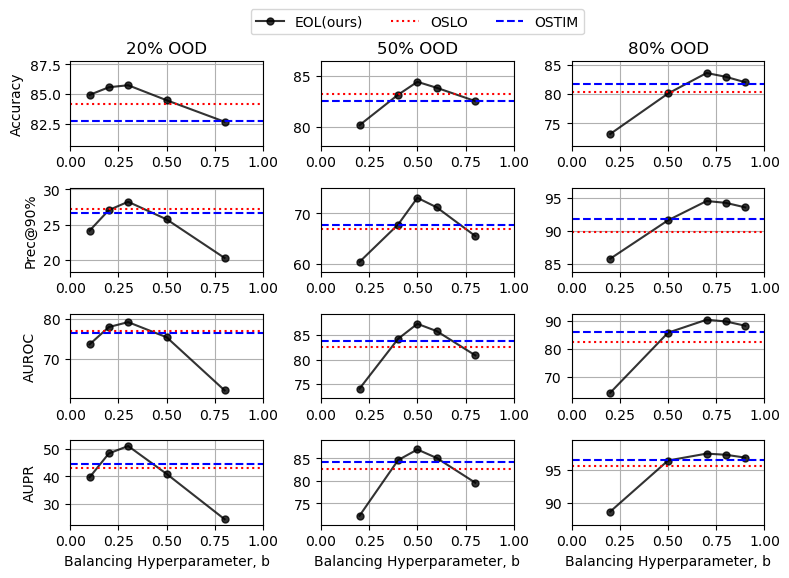}
    \small
    \caption[Study of the balancing hyperparameter]{\textbf{Study of the balancing hyperparameter, b.} We evaluate the performance of~\ourmethod~with different hyperparameters on different imbalance ratios. For comparison, we show the performance of two other transductive baselines, OSLO \cite{boudiaf2023osem}, and OSTIM \cite{boudiaf2022ostim} for each of the imbalanced settings.}
    \label{ch:openset:fig:hyperparam_b}
\end{figure}

\subsection{Ablation Study}

\subsubsection{Hyperparameter $b$.} 
Next, we consider the effect of hyperparameters on the performance of EOL in Fig.~\ref{ch:openset:fig:hyperparam_b}. Here, we investigate the effect of the balancing hyperparameter $b$ responsible for balancing the influence of inliers and outliers. In this empirical evaluation of $b$, \ourmethod's performance was systematically assessed across various imbalance ratios: 20\%, 50\%, and 80\% OOD. For comparison, we add the performances of two other baselines in the same category of methods: OSLO \cite{boudiaf2023osem} and OSTIM \cite{boudiaf2022ostim}. 

\textbf{Key Observations.} The highest achieving values of $b$ is correlated with the concentration of outliers in the outlier detection task. In the 20\% OOD scenario, $b \approx 0.3$ achieved peak accuracy, AUROC, AUPR, and Prec@0.9. As the imbalance ratio increased to 50\% and 80\% OOD, the optimal $b$ value for accuracy and AUROC shifted towards higher values, reflecting the necessity of a stronger balancing effect in the presence of greater outlier samples. In the 50\% and 80\% OOD settings, $b\approx0.5$ and $b\approx0.7$ achieved highest performance, respectively. The sensitivity of the $b$ suggests a delicate balance in achieving high-accuracy generalisation and outlier detection. 

\subsubsection{Learnable parameters, $\eta$ and $\delta$.}
Recall that we addressed the model calibration problem by introducing two sets of learnable parameters $\eta$ and $\delta$, defined in (\ref{ch:openset:eq:cosine_calib}), which are optimised through gradient descent during transductive inference. We consider the effect of these parameters on the performance of the EOL algorithm. We conduct a systematic study across varying imbalance ratios (20\%, 50\%, and 80\% OOD). The investigation, as summarised in Table~\ref{ch:openset:tab:hyperparameter etadelta}, considered scenarios with and without applying these parameters, examining their influence on the previously introduced metrics.

\begin{table}[tbh]
    \centering
    \small
    \caption[Study of $\eta$ and $\delta$ under imbalance]{Study of $\eta$ and $\delta$ under imbalance.}
    \label{ch:openset:tab:hyperparameter etadelta}
    \scalebox{0.75}{
    \begin{tabular}{c|cccc|cccc|cccc}
    \toprule
    \%OOD & \multicolumn{4}{c|}{20\%} & \multicolumn{4}{c|}{50\%} & \multicolumn{4}{c}{80\%} \\
    Hyperparams &   Acc. &   AUPR & Prec@0.9 &  AUROC &   Acc. &   AUPR & Prec@0.9 &  AUROC &   Acc. &   AUPR & Prec@0.9 &  AUROC \\
    \midrule
    no $\eta$, no $\delta$ &  84.48 &  43.29 &    25.11 &  74.67 &  84.31 &  86.58 &    72.33 &  86.95 &  82.31 &  95.73 &    91.64 &  84.09 \\
    $\eta$               &  84.91 &  41.88 &    24.87 &  74.70 &  84.37 &  86.15 &    72.30 &  86.67 &  82.36 &  95.76 &    91.57 &  84.67 \\
    $\delta$             &  85.44 &  45.30 &    25.93 &  76.51 &  84.27 &  86.56 &    72.20 &  86.86 &  82.84 &  97.32 &    94.10 &  89.64 \\
    $\eta$, $\delta$       &  85.58 &  48.31 &    27.06 &  77.88 &  84.39 &  86.97 &    73.10 &  87.35 &  83.00 &  97.34 &    94.20 &  89.75 \\
    \bottomrule
    \end{tabular}
}\end{table}%

\textbf{Key Observations.} 
Notably, including both $\eta$ and $\delta$ parameters consistently improve performance across all evaluated metrics and settings by $+0.6$, $+2.3$, $+1.8$, $+3.1$ point advantage in accuracy, AUPR, Prec@0.9, and AUROC, respectively. Among the two parameters, $\delta$ is the more influential parameter, contributing substantially to the performance uplift. This is evident from the significant increments in AUPR and Prec@0.9 when $\delta$ is applied, either individually or in conjunction with $\eta$. Interestingly, the solitary inclusion of \( \eta \) displays a nuanced behaviour; it has no effect on performance, and in scenarios with a lower proportion of OOD samples, it even slightly detracts from performance. This suggests that \( \eta \), while beneficial in tandem with \( \delta \), may not always be beneficial. Conversely, the joint application of both \( \eta \) and \( \delta \) proves to be mutually advantageous, outstripping the benefits of deploying \( \delta \) alone. This suggests the necessity of a balanced hyperparameter strategy, one that harnesses the strengths of both \( \eta \) and \( \delta \) to achieve optimal model performance across the spectrum of inlier-outlier imbalance settings.

\section{Conclusion}
In summary, in this work, we explored the more nuanced and practical challenge of the Open-Set Few-Shot Learning (OSFSL) setting that incorporates unknown classes into the query set, thereby requiring the model not only to classify known classes but also to identify outliers. Our work presented a novel method called EOL. Through an empirical evaluation, we demonstrated that our approach was able to achieve superior performance compared to other methods in a variety of tasks. This suggests that appropriately focusing on model calibration and balancing inlier-outlier ratios can significantly improve the evaluation performance. Our proposed solution recorded performance improvements over the baseline OSTIM method, ranging from approximately $+1.3\%$ to $+6.3\%$ across balanced and imbalanced query settings. Future work could investigate the online estimation of the $b$ hyperparameter, allowing the model to have reduced hyperparameter dependency. 

\clearpage  

%
%
\bibliographystyle{splncs04}
\bibliography{egbib}
\end{document}